\documentclass{article} 
\usepackage{iclr2023_conference,times}

\usepackage{amsmath,amsfonts,bm}

\def\eqref#1{equation~\ref{#1}}

\def\1{\bm{1}}

\DeclareMathAlphabet{\mathsfit}{\encodingdefault}{\sfdefault}{m}{sl}
\SetMathAlphabet{\mathsfit}{bold}{\encodingdefault}{\sfdefault}{bx}{n}

\usepackage{hyperref}
\usepackage{url}
\usepackage{graphicx}
\usepackage{booktabs}
\usepackage[ruled,linesnumbered]{algorithm2e}
\title{Meta-Ensemble Parameter Learning}

\author{Zhengcong Fei, Shuman Tian, Junshi Huang\thanks{Corresponding author.}, Xiaoming Wei, Xiaolin Wei \\
Meituan\\
Beijing, China\\
\texttt{\{name\}@meituan.com} \\
}

\iclrfinalcopy 
\begin{document}

\maketitle

\begin{abstract}
Ensemble of machine learning models yields improved performance as well as robustness. However, their memory requirements and inference costs can be prohibitively high. Knowledge distillation is an approach that allows a single model to efficiently capture the approximate performance of an ensemble while showing poor scalability as demand for re-training when introducing new teacher models. In this paper, we study if we can utilize the meta-learning strategy to directly predict the parameters of a single model with comparable performance of an ensemble. Hereto, we introduce WeightFormer, a Transformer-based model that can predict student network weights layer by layer in a forward pass, according to the teacher model parameters. The proprieties of WeightFormer are investigated on the CIFAR-10, CIFAR-100, and ImageNet datasets for model structures of VGGNet-11, ResNet-50, and 
ViT-B/32, where it demonstrates that our method can achieve approximate classification performance of an ensemble and outperforms both the single network and standard knowledge distillation. More encouragingly, we show that WeightFormer results can further \textbf{exceeds average ensemble} with minor fine-tuning. Importantly, our task along with the model and results can potentially lead to a new, more efficient, and scalable paradigm of ensemble networks parameter learning.  

\end{abstract}

\section{Introduction} 

As machine learning models are being deployed ever more widely in practice, memory cost and inference efficiency become increasingly important \citep{bucilua2006model,polino2018model}. Ensemble methods, which train several independent models to form a decision, are well known to yield both improved performance and reliable estimations \citep{perrone1992networks,drucker1994boosting,opitz1999popular,dietterich2000ensemble,sagi2018ensemble}. Despite their useful property, using ensembles can be computationally prohibitive. Obtaining predictions in real-time applications is often expensive even for a single model, and the hardware requirements for serving an ensemble scales linearly with number of teacher models \citep{buizza1998impact,bonab2019less}. As a result, over the past several years the area of knowledge distillation has gained increasing attention \citep{hinton2015distilling,freitag2017ensemble,malinin2019ensemble,lin2020ensemble,park2021learning,zhao2022decoupled}. Broadly speaking, distillation methods aim to involve a single student model which can approximate the behavior of a teacher ensemble, but at a low computational cost. In the simplest and most frequently used form of distillation \citep{hinton2015distilling}, the student model is trained to capture the average prediction of the ensemble,  
\emph{e.g.}, in the case of image classification, this reduces to minimizing the KL divergence between the soft labels of student model and teacher models.

\begin{figure*}[t]
	\centering
	\includegraphics[width=1.0\columnwidth]{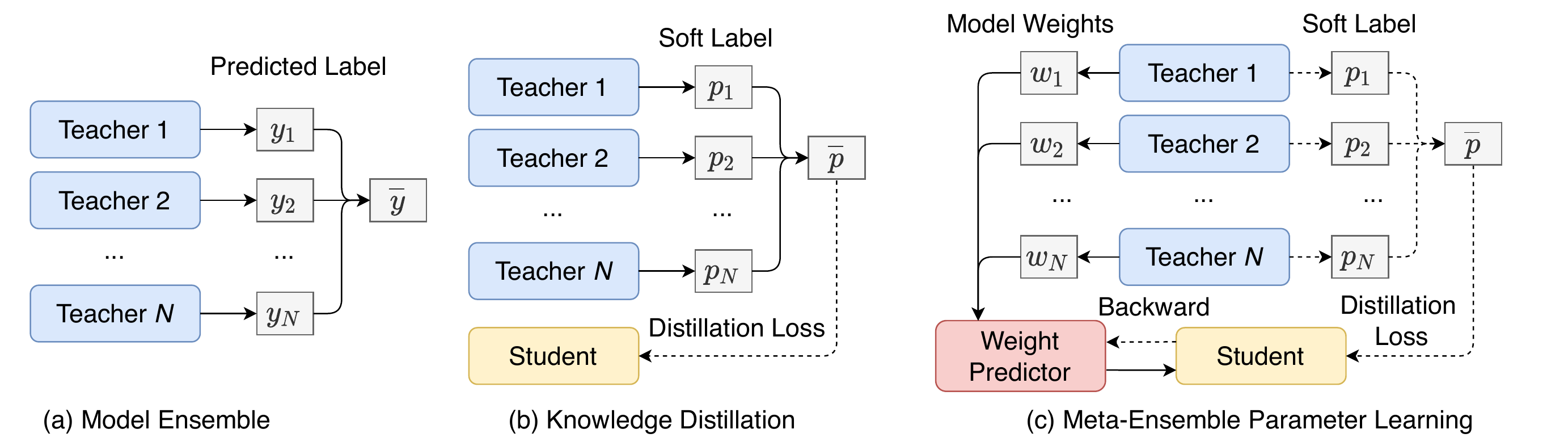}
	\caption{Illustration of different knowledge induction frameworks. }
	\label{fig:flow}
\end{figure*}

When optimizing the parameters for a new ensemble model, typical knowledge distillation process disregards information on teacher model parameters and past training experience for distillation of teacher models. However, leveraging this training information can be the key to reduce the high computational demands. 
To progress in this direction, we propose a new task, referred to as \emph{Meta-Ensemble Parameter Learning}, where parameters of the distillation student model are directly predicted with a weight prediction network. The main idea is to use deep learning models to learn the parameter distillation process and finally generate an entire student model by producing all model weights in a single pass. This can reduce the overall computation cost in cases where the tasks or ensemble models update frequently. It is important to highlight that meta-ensemble parameter learning, to our knowledge, has not been previously investigated. Figure \ref{fig:flow} depicts various information transfer paradigms, including model ensemble, knowledge distillation, and meta-ensemble parameter learning. The dotted line represents the training flow.

To cover this task, we introduce WeightFormer, a model to directly predict the distilled student model parameters. Our architecture takes inspiration from the Transformer \citep{attention} and incorporates two key novelties to imitate the characteristics of model ensemble, \emph{i.e.}, cross-layer information flow and shift consistency constraint. By designing these updated techniques, we then evaluate the classification performance obtained by the predicted parameters on conventional convolutional architectures VGGNet-11, ResNet-50, and transformer architecture ViT-B/32 \citep{dosovitskiy2020image}, on the CIFAR-10, CIFAR-100, and ImageNet datasets, respectively. Experimental results show that the predicted models, yielded by our proposed method in one forward pass, approach the performance of average ensemble and outperform the regular knowledge distillation models significantly. 
Besides, our WeightFormer further exceeds the average ensemble with fine-tuning, which is expected to hold good fitness on the variability and complexity of application scenarios.

Overall, our framework and results pave the road toward a new and significantly more efficient paradigm for ensemble model parameter learning. The contributions of this paper are  summarized as follows: 
$i$) We introduce a novel task of directly predicting parameters of the distillation student model based on multiple teacher neural network parameters, which encourages the exploring of past ensemble training experience to improve the performance as well as reduce computation demand; $ii$) We design WeightFormer, a simple and effective benchmark, with adjustments of built cross-layer information flow and shift consistency constraints, to track progress on the model weight generation task. Experimentally, our approach performs surprisingly well and robustly on different model architectures and datasets; $iii$) We show that WeightFormer can be transferred to the scenario of weight generation for \textbf{unseen teacher models} in a single forward pass, and \textbf{more competitive results} can be obtained with additional fine-tuning data. Moreover, to improve the reproducibility and foster new research in this field, we will publicly release the source code and trained models. 



\section{Task Formulation}

Here we consider the problem of distilling a neural network from several trained neural networks, also known as teacher-student paradigm \citep{hinton2015distilling}, on image classification task \citep{rokach2010ensemble}. It essentially aims to train a single student model that capture the mean decision of an ensemble, allowing to achieve a higher performance with a far lower computation cost. This problem can be formalized as finding optimal parameters $\widetilde{w}$ for target neural network $\widetilde{f}$, given a neural network set $\mathcal{F}=\{f_1, \ldots, f_N \}$ parameterized by $\mathcal{W} = \{w_1, \ldots, w_N\}$, w.r.t. a loss function on the dataset $\mathcal{D}=\{ (x_i, y_i)\}_{i=1}^M$ of input image $x_i$ and ground truth label $y_i$:
\begin{equation}
    \begin{aligned}
        \min_{\widetilde{w}} \sum_{i=1}^M \text{KL}(\widetilde{p}(x_i) || \frac{1}{N}\sum_{n=1}^N p_n(x_i)),
        \label{eq:1}
    \end{aligned}
\end{equation}
where the optimization objective includes  Kullback-Leible divergence denoted as KL($\cdot$) between the mean soft labels from teacher models and the predictions from student model. $p_n(\cdot)$ is the output distribution of $n$-th network and $\widetilde{w}$ is the resulting parameters of ensemble distillation model.
Here we assume all the teacher models hold the same network architecture and leave the ensemble learning of heterogeneous models as future work. Despite the progress in memory saving for distillation ensemble model $\widetilde{f}$, obtaining
$\widetilde{w}$ remains a bottleneck in large-scale machine learning pipelines. In particular, with the growing size of network, the classical process of obtaining ensemble parameters, retraining from scratch, is becoming computationally unsustainable.

In this paper, we highlight that the knowledge of preceding ensemble training in parameter optimization is also important and propose a new task, named \emph{Meta-Ensemble Parameter Learning}, where parameters of distillation ensemble model are directly predicted with deep learning network. Formally, the task aims to generate the parameter $\widetilde{w}$ of target model $\widetilde{f}$ in a single forward pass using a specific weight generation network $g_\theta$, parameterized by $\theta$: 
\begin{equation}
    \widetilde{w} = g([w_1, \ldots, w_N]; \theta). \label{eq:2}
\end{equation}
This task is constrained to a dataset $\mathcal{D}$, so $\widetilde{w}$ is the predicted parameter for which the test set performance of $\widetilde{f}(x; \widetilde{w})$ is approximate to the performance of model ensemble while maintaining a training efficiency and scalability. In this manner, we can even distill the unseen teacher models to achieve competitive performance without any training cost.

\section{Methodology}

\begin{figure*}[t]
	\centering
	\includegraphics[width=1.0\columnwidth]{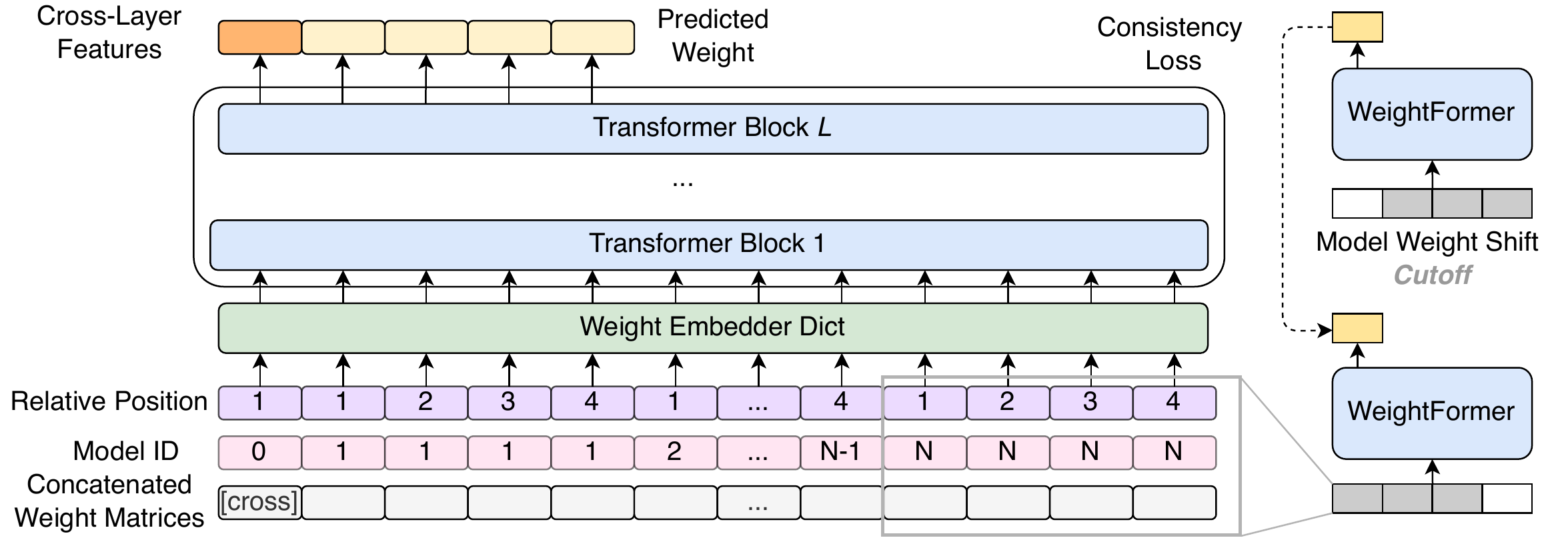}
	\caption{Overview of WeightFormer for the generation of one layer weights. Transformer-based weight generator receives concatenated weight matrices of teacher models along with model id and position information and produces the corresponding layer weights. After being generated, the predicted student model is used to compute the loss on the training set, whose gradients are then used to update the weights of WeightFormer. “[cross]” is a special token placed at the beginning of all weight matrices to model the cross-layer information flow. The right part illustrates the process of shift consistency, where predicted layer weights should be consistent with shifted input models (see light brown token).}
	\label{fig:framework}
\end{figure*}

In this section, we will describe our approach, dubbed as WeightFormer, to serve as an effective solution for meta-ensemble parameter learning based on the Transformer structure. For simplicity, we describe the prediction of CNN models containing a set of convolutional layers and two fully-connected logits layers as well as self-attention layers in Transformer.   Please note that most of common parametric layers can be predicted by WeightFormer as presented in experiments.

\subsection{Representation of Model Parameters}

For weight matrices in different layers of teacher models, provided with the convolutional kernel size $k$ and input / output channel number $n_{input}$ / $n_{output}$, we consider the encoding of $k \times k \times n_{input} \times n_{output}$ convolutional kernels as $n_{output}$ tokens with weight slices of $k^2 \times n_{input}$ dimensionality and fully-connected logits layer $n_{input} \times n_{output}$ weights as $n_{output}$ tokens with dimensionality of $n_{input}$ \citep{zhmoginov2022}.  For parameters of self-attention layer, which includes $h$ projection query $W^Q \in \mathbb{R}^{{d}_{trans}\times d_k}$, key $W^K\in \mathbb{R}^{{d}_{trans}\times d_k}$, value $W^V\in \mathbb{R}^{{d}_{trans}\times d_v}$, and one concatenation matrix $W^O \in \mathbb{R}^{h d_v \times {d}_{trans}}$, the weight sequence is modeled as $2d_k + 2d_v$ length and $h \times {d}_{trans}$ dimension. 
The concatenated weight sequence of different models is then fed into weight embedding dictionary layer by layer to achieve identical dimensional as WeightFormer model $d_{model}$. The weight embedding dictionary $\mathcal{D}_w: \mathbb{R}^{d_{layer}} \rightarrow \mathbb{R}^{d_{model}} \text{ and }  \mathbb{R}^{d_{model}} \rightarrow \mathbb{R}^{d_{layer}}$ includes a cluster of dimension adjustment embedding functions for different network layers, where $d_{layer}$ denotes the dimension of original weight token. 
To be specific, it normalize the input weight dimension into $d_{model}$ before WeightFormer and revert the $d_{model}$ into output weight dimension after re-mapped with $\mathcal{D}_w$. So that, we can easily extend our WeightFormer for variant layers with a shared weight generator in a \emph{\textbf{unified way}}.

To make our WeightFormer model obtaining the ability to distinguish among the different parts of input weights and making use of the order of sequence, the final representation for each weight token is obtained by summing up its weight feature embedding, position embedding, and model id embedding, as illustrated in Figure \ref{fig:framework}. Note that we use relative position to distinguish the parameters of output dimension in one model. 


\subsection{Design of Architecture}

As introduced in the preceding discussion, modeling of parameter predictor $g$ is the core of the parameter learning task and here we choose $g$ to be a model of stacked Transformer blocks that takes the parameter of teacher models $\{w_1, \ldots, w_N \}$ as input and produces weights for some or all layers $\widetilde{w}$ of the target ensemble model. 
Generally, the weights are generated layer-by-layer from the first layer.
Next, we will introduce the key novelties in detail.

\paragraph{Cross-Layer Information Flow.} 
When the WeightFormer predicts the parameters of specific layer, the cross-layer information should be incorporated as prior context for holistic parameters prediction, as the current layer is not independent with the previous layers.
To achieve this goal, we extend the original weight matrices sequence with an extra vector [cross], which will be inserted at the first position and fully attended with other weight tokens in a global perspective. The corresponding output hidden states, which contain the layer information and serve as cross-layer features, are fed into the parameter prediction process of the next generated layer.
The value of [cross] token in the generation of the first layer is a randomly initialized learnable vector.
Particularly, we claim that the output hidden states implicitly model the embedding of current layer index, and is critical for layer-specific generation.



\paragraph{Shift Consistency.}

Motivated by the fact that for the same teacher models with different input orders, the outputs of ensemble parameter prediction should be identical. 
Hereto, we develop a shift consistency loss to regularize the parameter prediction with shifted input sequences.
As shown in the right part of Figure \ref{fig:framework}, we introduce a model weight shifted topology. Concretely, given the input weight sequence $\mathcal{W}$ at each training step, we feed the same input $\mathcal{W}$ to go through the forward pass of the WeightFormer network twice, one with the original model weight order, and the other with the one-model shifted order. For these two patterns, we use different weight cutoff \citep{budczies2012cutoff}, where the dropped values are chosen from feature dimension of $d_{model}$. Therefore, we can obtain two parameter predictions $\widetilde{w}$ and $\widetilde{w}^{shift}$, respectively.
In order to make those two predicted weights closing to each other, we propose to minimize the mean square error between them as: 
\begin{equation}
    \mathcal{L}_{consist} =||\widetilde{w} - \widetilde{w}^{shift}||^2 . \label{eq:4}
\end{equation}
Intuitively,  the regularization naturally forces the outputs of  WeightFormer with different input orders of teacher models to be identical and invariant. 
After the entire input weight sequence is processed by WeightFormer, we extract the  first $n_{output}$ tokens starting from [cross] and assemble the final student model by adjusting the weights' dimension with layer-specific embedding dictionary.

\subsection{Training Procedure}

Instead of direct learning for student model parameters, we follow the bi-level optimization paradigm common in meta-learning \citep{hospedales2020meta,liu2021investigating} to train the WeightFormer model as: 
\begin{equation}
    \mathcal{L}_{comb} =  \sum_{(x_i, y_i) \in \mathcal{D}} \mathcal{L}_{cross}(f_t(x_i; \widetilde{w}), y_i) + \alpha \mathcal{L}_{consist}, \label{eq:3}
\end{equation}
where $\mathcal{L}_{cross}$ is the single-label classification cross-entropy loss and $\alpha$ denotes the balancing factors for consistency loss. By optimizing the joint loss, the WeightFormer $g$ gradually obtains the capacity of how to capture the intrinsic knowledge of network parameters from teacher models.


\begin{minipage}[t]{0.5\linewidth}
During training, the WeightFormer model inputs the combination of the teacher models' parameters layer by layer to produce the weights of all layers of student model. Then, the loss in Equation \ref{eq:3} is computed for the training set samples that are passed through the generated student model. 
The weight generation parameters $\theta$ of WeightFormer are learned by optimizing the loss function using stochastic gradient descent. 
To avoid too much costs on reloading parameter at each step, we re-sample the teacher model checkpoints from model set only meets the break conditions, \emph{e.g.}, reaching every predefined steps or converging on validation set. 
The overall algorithm is presented in Algorithm \ref{algorithm}. 
Encouragingly, our WeightFormer shows a good generalization and predicts good parameters for different architectures, even for unseen teacher models. 
\end{minipage}
\hfill
\begin{minipage}[t]{0.43\linewidth}
\small
\begin{algorithm}[H]
\caption{\small{Optimizing for WeightFormer model g$_\theta$}}\label{algorithm}
\KwData{Training data $\mathcal{D}_{train}$, validation data $\mathcal{D}_{val}$, model set $\mathcal{F}$, model ensemble number $n$}
\For{ sample $n$ models $ \in \mathcal{F}$}{
    \For{sample data pair $(x_i, y_i)$ $ \in \mathcal{D}$}{
        $\widetilde{w} = g_{\theta}(w_1, \ldots, w_n)$\;
        compute loss $\mathcal{L}_{comb}$\; 
        loss backward for $\theta$\;
        \If{stop condition $(\widetilde{w}, \mathcal{D}_{val})$}
        {break\;}
    }
}
\end{algorithm}

\end{minipage}


\section{Experiments} 

This section focuses on the evaluation of WeightFormer on the ensemble parameter learning task. 
We begin with experimental setup in Sec. \ref{sec:exp_setting} and compare our main results with various baselines in Sec. \ref{sec:comparison}. In between, we show beneficial side-effects of predicting ensemble weights and discuss the implications of our findings. 
Finally, we conduct more analysis on the design of WeightFormer in Sec. \ref{sec:model_analysis} for empirical study.

\subsection{Experimental Settings}
\label{sec:exp_setting}

\paragraph{Datasets and Metrics.} We utilize three common image classification datasets CIFAR-10 (C-10), CIFAR-100 (C-100) \citep{krizhevsky2009learning}, and ImageNet \citep{deng2009imagenet} for experiments. 
To evaluate the image classification performance, we adopt the metrics as 1) top-$N$ accuracy (ACC-$n$), which means any of model's top $n$ highest probability predictions match with the expected results; 2) expected calibration error \citep{guo2017calibration} (ECE), which simply takes a weighted average over the absolute accuracy / confidence difference.

\paragraph{Compared Baselines.} 
The following baselines for knowledge induction are considered: 
1) \emph{Single} refers to the average performance across different individual, independently evaluated models. 
2) \emph{Ensemble} refers to the performance of a deep ensemble that average the sum of logits from ensemble models, which is currently considered to be both the best and the simplest ensemble method.
3) \emph{KD} refers to ensemble distillation via minimizing the KL divergence between the student and the ensemble’s mean for image classification. In practice, KD is executed with a fixed temperature and same structure to teacher network following \citep{hinton2015distilling}.
4) \emph{MLP} denotes a simple MLP that only has access to weight parameters of teacher models to predict the target ensemble parameter, but not to the connections between different layers. Note that the prediction of parameters for each layer requires an independent MLP network.

\paragraph{Implementation Details.} 
WeightFormer can in principle generate arbitrarily large weight tensors by concatenating all the parameters of teacher models as input and then predicting the student model weight tensors layer by layer. 
For model architecture, the proposed WeightFormer model closely follows the same network architecture and hyper-parameters settings as an encoder version of Transformer model \citep{attention}. Specifically, the number of stacked blocks for the transformer block encoder is set to 24, hidden size is set to 1280, attention head number is set to 20, and feed-forward network size is set to 4096. For training process, to avoid the collapse, we first pre-training the WeightFormer with the supervised of knowledge distillation model parameters in L2 matching loss using 1e-3 SGD. Then, we first sample checkpoints from model set and data pair from training set, and then train the WeightFormer model in combined loss with an initial learning rate of 3e-5, and it decays by 0.9 every five epochs. 
To reduce the cost of parameter loading at each step, we reload or sample model checkpoints every 5k steps. 
The overall training process is terminated until accuracy achieves no increment in the validation set. 
The hyper-parameter factors $\alpha$ is set to 1.0 according to prior experiments.

\subsection{Compare with knowledge Induction Methods} 
\label{sec:comparison}

We first turn our attention to ensemble model parameter prediction with different knowledge in methods. This section presents our main results, which shows that proposed WeightFormer models can be served as a competitive alternative to the conventional procedure of ensemble methods which require no additional training or memory cost.

\paragraph{Experimental Setup.}

To evaluate the performance of ensemble parameter learning, a model set including 72 checkponits is constructed by training with different random seed initializations and hyperparameters \emph{dependently}, for VGGNet-11, ResNet-50, and ViT-B/32 versions on the C-10, C-100, and ImageNet datasets. In each setting, 60 models and 12 models is splitted for training and evaluation. 
Checkpoints of ViT-B/32 are loaded from repository \footnote{https://github.com/mlfoundations/model-soups} and please refer to \citep{wortsman2022model} for the detail settings of hyperparameter. 
As the original downloaded model of ViT-B/32 fine-tuning on the ImageNet, we only conduct experiments on the ImageNet dataset for performance comparison. 
We consider the scenario of 3 model ensemble, \textit{i.e.}, 3 checkpoints are randomly sampled from model pools of training and evaluation, respectively. The checkpoints for evaluation is identical for different methods for fair comparison. Except for the four baselines introduced in section 4.1, we also use WF*, which denotes fine-tunning the weightformer with sampled unseen checkpoints in the corresponding training dataset until convergence in the validation set. 

\begin{table}[t]
	\begin{center}
		\setlength{\tabcolsep}{2mm}{
			\begin{tabular}{llcccccc}
				Dataset& Metrics&Single&Ensemble&KD&MLP&WF&WF*\\
			\midrule
			&ACC-1&92.4 \tiny${\pm}$0.6& {93.8}  \tiny${\pm}$NA&92.8 \tiny${\pm}$0.4&92.5 \tiny${\pm}$0.6&93.3 \tiny${\pm}$0.3 & \textbf{93.9}  \tiny${\pm}$0.2 \\
		C-10&ACC-5&98.8 \tiny${\pm}$0.2& \textbf{99.8} \tiny${\pm}$NA &99.1 \tiny${\pm}$0.1&98.8 \tiny${\pm}$0.4&99.3 \tiny${\pm}$0.1 &\textbf{99.8} \tiny${\pm}$0.1\\
			&ECE&2.5 \tiny${\pm}$0.6&{1.1} \tiny${\pm}$NA&2.3 \tiny${\pm}$0.5&2.5 \tiny${\pm}$0.5&1.4 \tiny${\pm}$0.3 &\textbf{1.0}\tiny${\pm}$0.2\\
			\midrule
			&ACC-1&69.8 \tiny$\pm$0.5&{72.6} \tiny$\pm$NA&71.3 \tiny$\pm$0.3&70.0 \tiny$\pm$0.6&72.0 \tiny$\pm$0.3 &\textbf{72.9} \tiny$\pm$0.3 \\
		C-100&ACC-5&90.5 \tiny$\pm$0.3& {92.4} \tiny$\pm$NA&91.0 \tiny$\pm$0.1&90.7 \tiny$\pm$ 0.2&91.6 \tiny $\pm$0.1  &\textbf{92.5} \tiny$\pm$0.2\\
			&ECE&10.1 \tiny$\pm$0.5&{3.2} \tiny$\pm$NA&7.7 \tiny$\pm$0.4&9.8 \tiny $\pm$0.5&5.4 \tiny $\pm$0.3 &\textbf{3.0} \tiny$\pm$0.3 \\
			\midrule
			&ACC-1&70.3\tiny$\pm$0.5&\text{73.4}\tiny$\pm$NA&71.1\tiny$\pm$0.2&70.5\tiny$\pm$0.8&72.2\tiny$\pm$0.5 &\textbf{73.6}\tiny$\pm$0.4\\
	ImageNet&ACC-5&89.8\tiny$\pm$0.4&\textbf{91.4}\tiny$\pm$NA&90.3\tiny$\pm$0.2&89.9\tiny$\pm$0.2&90.9\tiny$\pm$0.1&91.3\tiny$\pm$0.2\\
			&ECE&19.6\tiny$\pm$0.8&{5.5}\tiny$\pm$NA&9.2\tiny$\pm$0.3&18.5\tiny$\pm$0.7&6.3\tiny$\pm$0.5&\textbf{5.2}\tiny$\pm$0.3\\
			\bottomrule
			\end{tabular}
			{\caption{ Experimental results on the C-10 / C-100 / ImageNet classification datasets across three teacher models of VGGNet-11 architecture $\pm 2 \sigma$.}\label{tab:1}}
		}
	\end{center}
\end{table}

\begin{table}[t]
	\begin{center}
		\setlength{\tabcolsep}{2mm}{
			\begin{tabular}{llcccccc}
				Dataset& Metrics&Single&Ensemble&KD&MLP&WF&WF*\\
			\midrule
			&ACC-1&93.6 \tiny${\pm}$0.7& \text{94.7}  \tiny${\pm}$NA&93.9 \tiny${\pm}$0.4&93.5 \tiny${\pm}$0.5&94.3 \tiny${\pm}$0.6 & \textbf{94.8}  \tiny${\pm}$0.3\\
	C-10&ACC-5&99.7 \tiny${\pm}$0.1& \textbf{99.8} \tiny${\pm}$NA &99.7 \tiny${\pm}$0.1&99.7 \tiny${\pm}$0.1&\textbf{99.8} \tiny${\pm}$0.1&\textbf{99.8} \tiny${\pm}$0.05\\
			&ECE&2.1 \tiny${\pm}$0.5&\textbf{0.9} \tiny${\pm}$NA&1.8 \tiny${\pm}$0.5&2.3 \tiny${\pm}$0.5&1.6 \tiny${\pm}$0.4 & \text{1.0} \tiny${\pm}$0.2\\
			\midrule
			&ACC-1&77.4 \tiny$\pm$0.8&\text{79.7} \tiny$\pm$NA&78.2 \tiny$\pm$0.4&77.6 \tiny$\pm$0.4&78.9 \tiny$\pm$0.4 &\textbf{79.9} \tiny$\pm$0.4\\
	C-100&ACC-5&93.9 \tiny$\pm$0.2& \text{94.8} \tiny$\pm$NA&94.1 \tiny$\pm$0.2&94.0 \tiny$\pm$ 0.1&94.5 \tiny $\pm$0.1 & \textbf{94.9} \tiny$\pm$0.1\\
			&ECE&10.4 \tiny$\pm$0.6&\text{3.0} \tiny$\pm$NA&6.3 \tiny$\pm$0.6&9.3 \tiny $\pm$0.5&4.3 \tiny $\pm$0.4 &\textbf{2.7} \tiny$\pm$0.2\\
			\midrule
			&ACC-1&76.1\tiny$\pm$0.6&\text{78.3}\tiny$\pm$NA&77.1\tiny$\pm$0.3&76.2\tiny$\pm$0.3&77.9\tiny$\pm$0.3 &\textbf{78.7}\tiny$\pm$0.5\\
	ImageNet&ACC-5&92.8\tiny$\pm$0.3&\text{94.1}\tiny$\pm$NA&93.2\tiny$\pm$0.2&92.8\tiny$\pm$0.1&93.8\tiny$\pm$0.1&\textbf{94.3}\tiny$\pm$0.2\\
			&ECE&16.5\tiny$\pm$1.0&\text{4.6}\tiny$\pm$NA&10.3\tiny$\pm$0.8&16.8\tiny$\pm$0.6&6.5\tiny$\pm$0.5& \textbf{4.5}\tiny$\pm$0.2\\
			\bottomrule
			\end{tabular}
			}
			{\caption{ Experimental results on C-10 / C-100 / ImageNet classification datasets across three teacher models of  ResNet-50 architecture $\pm 2 \sigma$.}\label{tab:2} }
	\end{center}
\end{table}


\paragraph{Main Results.} 

Table \ref{tab:1}, \ref{tab:2} and \ref{tab:3} report the evaluation results of our ensemble parameter distillation experiments for network architectures of VGGNet-11, ResNet-50, and ViT-B/32 on the C-10, C-100, and ImageNet datasets. We can see that 1) The ensemble approach significantly outperforms single model. This is canonical approach and therefore ensemble strategy is generally adopted when the absolute best score is desired. 
2) Another interesting result is that we are able to come rather close to the performance of the ensemble with our WeightFormer approach for unseen model checkpoints. As a result, to fit different teacher models, KD would require more training costs and be limited by the model scalability. Given that our goal is to obtain the performance of ensemble in one feed forward pass without retraining, we believe that this result is considerably successful. 3) Our WeightFormer outperforms the single model baseline by an average of 1.7\% improvements on ACC-1 metric on the ImageNet dataset and similar trends as on the other two datasets. Meantime, for ECE, which compares neural network model output pseudo-probabilities to model accuracy, the classification performance changes are more evident.
4) Results from WeightFormer are shown with less variance and more robust prediction performance compared with KD, which illustrates more superiority of the proposed WeightFormer.
5) Simple MLP baseline shows a much poor transfer capacity for unseen checkpoints, while Transformer with well designed improvements hold more promising. 
6) More encouraging, fine-tuned weightformer with training data and teacher model parameters, the finally predicted model achieves better performance compared with average ensemble method in some metrics, \emph{e.g.}, 0.3\% ACC-1 improvement in ViT-B/32 architecture of ImageNet dataset, demonstrating the promising of meta ensemble paradigm.

\begin{table}
	\begin{center}
		\setlength{\tabcolsep}{2mm}{
			\begin{tabular}{llcccccc}
				Dataset& Metrics&Single&Ensemble&KD&MLP&WF&WF*\\
			\midrule
			&ACC-1&78.3\tiny$\pm$1.2&{80.2}\tiny$\pm$NA&79.0\tiny$\pm$0.8&78.4\tiny$\pm$0.5&79.8\tiny$\pm$0.4&\textbf{80.5}\tiny$\pm$0.3\\
	ImageNet&ACC-5&93.5\tiny$\pm$0.3&{94.9}\tiny$\pm$NA&94.0\tiny$\pm$0.2&93.9\tiny$\pm$0.1&94.8\tiny$\pm$0.2&\textbf{95.1}\tiny$\pm$0.2\\
			&ECE&2.6\tiny$\pm$0.4&{1.9}\tiny$\pm$NA&2.3\tiny$\pm$0.3&2.6\tiny$\pm$0.3&2.5\tiny$\pm$0.4& \textbf{1.7}\tiny$\pm$0.3\\
			\bottomrule
			\end{tabular}
			}
			{\caption{ Experimental results on the ImageNet classification dataset across three teacher models of  ViT-B/32 architecture $\pm 2 \sigma$.}\label{tab:3} }
	\end{center}
\end{table}

\subsection{Model Analysis} 
\label{sec:model_analysis}

In this section, we conduct analysis from different perspectives to better understand the advantages and properties of proposed WeightFormer, for the ensemble teacher models of VGGNet-11 / ViT-B on both C-10, C-100, and ImageNet image classification datasets.

\paragraph{Impact of Different Model Components.}

We investigate the impact of the different components in our proposed WeightFormer framework including cross-layer fusion and shift consistency (whether set the factor $\alpha$ in Equation 4 to zero), in the ensemble parameter learning task. To discriminate the contribution of consistency loss or used weight cutoff, we also test the baseline without cutoff augmentation. 
Table \ref{tab:4} shows the evaluation results of evaluation metrics on different variants. We can see that: 1) each component of the model contributes positively to its final classification performance and shift consistency yields the weight generation improvement most and 2) For different teacher model architectures, e.g., CNN or Transformer, different weightformer components hold different benefits. In particular, the more complex the teacher model structure, the higher the predicted ensemble revenue. For more advanced structure, \emph{e.g.}, Swin Transformer \citep{liu2021swin}, it is worth exploring in the future.

\begin{table}
	\begin{center}
		\setlength{\tabcolsep}{2mm}{
			\begin{tabular}{lccccccccc}
			Model &\multicolumn{4}{c}{VGGNet-11} & \multicolumn{2}{c}{ViT-B}\\
			\midrule
				Dataset	& \multicolumn{2}{c}{\text{C-10}} &\multicolumn{2}{c}{\text{C-100}}  &\multicolumn{2}{c}{\text{ImageNet}}\\ \midrule
				Model variant& ACC-1 &ACC-5& ACC-1 &ACC-5 & ACC-1 &ACC-5  \\ \midrule
				WeightFormer&\textbf{93.3} \tiny${\pm}$0.3& \textbf{99.3} \tiny${\pm}$0.1&\textbf{72.0} \tiny${\pm}$0.2& \textbf{91.6} \tiny${\pm}$0.1&\textbf{80.5} \tiny${\pm}$0.3& \textbf{95.1} \tiny${\pm}$0.2\\
				- Cross-layer Fusion&93.1 \tiny${\pm}$0.3&99.3 \tiny${\pm}$0.2&71.8\tiny${\pm}$0.3&91.3\tiny${\pm}$0.2&80.2\tiny${\pm}$0.4&95.0\tiny${\pm}$0.2\\
				- Shift Consistency&93.0 \tiny${\pm}$0.4&99.2 \tiny${\pm}$0.06&71.4\tiny${\pm}$0.5&91.1\tiny${\pm}$0.1&79.9\tiny${\pm}$0.5&94.8\tiny${\pm}$0.2\\ 
				- Weight Cutoff&93.0 \tiny${\pm}$0.3&99.2 \tiny${\pm}$0.1&71.6\tiny${\pm}$0.4&91.2\tiny${\pm}$0.1&80.1\tiny${\pm}$0.3&94.8\tiny${\pm}$0.1\\ 
				\bottomrule
			\end{tabular}
			}
			{\caption{Impact when changing different components in WeightFormer framework.}\label{tab:4}}
	\end{center}
\end{table}

\paragraph{Effect of Incorporated Ensemble Model Number.} 

\begin{figure*}
	\centering
	\includegraphics[width=1.0\columnwidth]{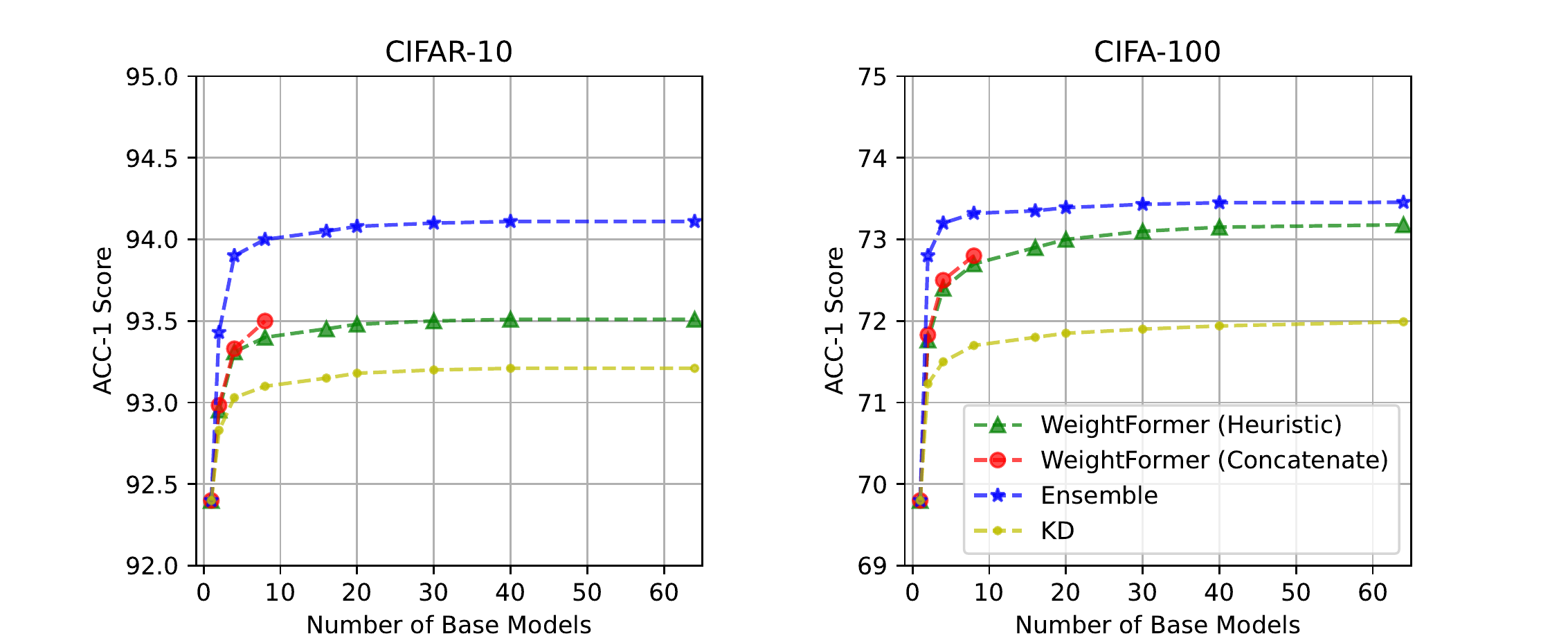}
	\caption{Classification performance when changing the incorporated teacher model number in different knowledge induction methods. }
	\label{fig:number}
\end{figure*}

Common sense shows that increasing the number of incorporated teacher models improves the final performance of the ensemble in an underlying task. It is natural to question whether our parameter prediction can benefit from more teacher model numbers. Hence, we conduct an ablation over the teacher model number using the \emph{\textbf{zero-shot}} VGGNet-11 model checkpoints over two configurations of our method: $i$) \emph{Heurastic}: select two teacher models without replacement each time for ensemble parameter prediction, and put the newly generated model into the teacher model set, and $ii$) \emph{Concatenate}: directly concatenate all teacher model parameters to predict the target weights. The results are summarized in Figure \ref{fig:number}. For each teacher model number and configuration, we directly predict the network weight with trained weightformer and report the metric scores over the test sets. Note that considering the memory constraints, the method of concatenation only reports part of the model number results.
As can be seen, increasing the teacher model number results in significant improvements in test evaluation. More importantly, the Transformer architecture allows increasing the incorporated teacher model number with only a marginal increment to number of learnable parameters.

\paragraph{Parameter Efficiency for WeightFormer.} 
Generally, our proposed approach learns a layer weight embedding per layer, consisting of $\sum_{l} n_{weight,l} \times d_{model}$ parameters, where $n_{weight,l}$ denotes the dimensional of input weight vector, \emph{i.e.}, $n_{input}\times k\times k$ for convolutional layer of kernel size $k$ and $n_{in}$ for full-linear layer and $d_{model}$ is the model dimensionality for Transformer block. We additionally employ model id and relative positional embeddings, which require $(N + L_{max}) \times d_{model}$ parameters, where $N$ is the teacher model number and $L_{max}$ is the max sequence length.  It is deserve to highlight that \emph{the structure of Transformer block in WeightFormer maintains the same in different ensemble scenarios as well as different layers}. In settings with a large number of layers and number of teacher model number $N$, our method shows much more parameter-efficient compared with MLP method. 



\section{Related Works}

\paragraph{Model Ensemble and Knowledge Distillation.} 

Combining the outputs of several independent models is a foundational technique for improving the accuracy and robustness of machine learning systems \citep{dietterich2000ensemble,bauer1999empirical,breiman1996bagging,lakshminarayanan2017simple,freund1997decision,ovadia2019can}.  \citep{gontijo2021no} conduct a large-scale study of ensembles, finding that higher divergence in training methodology leads to less uncorrelated errors and better ensemble accuracy. 
Meantime, plenty of works have explored building ensembles of models produced by hyperparameter searches \citep{mendoza2016towards,saikia2020optimized} and greedy selection strategies \citep{caruana2004ensemble,caruana2006getting,levesque2016bayesian,wenzel2020hyperparameter}. Importantly, ensembles require a separate inference pass through each model, which increases computational cost. When the number of models is large, this can be prohibitively expensive \citep{gou2021knowledge,wang2021knowledge}. 
Accordingly, knowledge distillation is proposed to allow the weak model (student) to learn existing knowledge from one-or-more strong models (teacher) by utilizing the output probability distributions as soft labels \citep{ba2014deep,hinton2015distilling}. 
Such a paradigm can greatly reduce the parameters of the model without reducing the performance of the model \citep{kim2016sequence}. \citep{freitag2017ensemble} transferred the knowledge of the ensemble model to a single model. 
Numerous works aim to refine and propagate core knowledge from the teacher networks, \emph{e.g.}, auto-encoders \citep{kim2018paraphrasing}, data instances relation \citep{park2019relational,tung2019similarity,peng2019correlation,liu2019knowledge}, mutual information \citep{ahn2019variational}, and contrastive learning \citep{tian2019contrastive,chen2021wasserstein}. However, all of the above knowledge distillation methods focused on the transfer of network output features, it is still a remaining problem if we can link knowledge with parameters between the teacher and student, which motivates our exploration work on ensemble parameter learning. 

\paragraph{Network Weight Modulation and Generation.}
The idea of using neural network and task-specification information to directly generate or modulate model weights has been previously explored \citep{bertinetto2016learning,ratzlaff2019hypergan,mahabadi2021parameter,tay2020hypergrid,ye2021learning}. Some few-shot learning methods described above employ this approach and use task-specific generation or modulation of the weights of the final classification model \citep{snell2017prototypical,lian2019towards,zhmoginov2022}. For example, the matching network in LGM-Net \citep{li2019lgm} is used to generate a few layers on top of a task-agnostic embedding. LEO \citep{rusu2018meta} utilized a similar weight generation method to generate initial model weights from the training dataset in a few-shot learning setting. These methods are tied to a particular architecture and need to be trained from scratch if it is changed. \citep{knyazev2021parameter,elsken2020meta} resorts to neural architecture search with graph network for more general parameter prediction. Most recently, \citep{wortsman2022model} propose model soup, to averages the weights of multiple models fine-tuned with different hyperparameter configurations often improves accuracy, which can be regarded as a special case of our parameter learning. 
Inspired by the approach \citep{ha2016hypernetworks}, we explore a similar idea, to directly generate an entire student model according to the parameters of teacher models, with simplifies and stabilizes training.

\section{Conclusion}
In this paper, we focus on ensemble parameter learning, which aims to directly generate the weights of distilled ensemble model. Accordingly, we introduce WeightFormer, a new Transformer architecture that leverages the representation of teacher learner parameters for distillation weight generation. 
The proposed approach is able to predict parameters for a distillation ensemble of diverse teacher models with cross-layer information flow modeling and shift consistency constraints.
The network with predicted parameters yields surprisingly high image classification accuracy given the challenging scenario of our task. More importantly, we showed that WeightFormer can be straightforwardly extended to handle unseen teacher models compared with knowledge distillation and even exceeds average ensemble with small-scale fine-tuning. 
We believe our task along with the model and results can potentially lead to a new paradigm of ensemble networks.

\paragraph{Broader Impact and Limitations.}  
Currently, to avoid the heavy memory cost in the model ensemble, advanced knowledge distillation methods are introduced. However, poor training extensibility restricts its application in more flexible scenarios. Our work, meta-ensemble parameter learning, takes the first step toward a general framework for direct parameter prediction from multiple teacher model parameters with conspicuous promise. However, on the other hand, one potential issue of this work is that it does not explore the weight generation for different teacher architectures. We advocate peer researchers look into this, making weight generation more reliable in different application scenarios and network architectures. 
Other than that, since this work is mostly on the discovery of ensemble weight generation, we do not foresee any direct negative impacts on society.

\bibliography{meta_ensemble}

\begin{thebibliography}{63}
\providecommand{\natexlab}[1]{#1}
\providecommand{\url}[1]{\texttt{#1}}
\expandafter\ifx\csname urlstyle\endcsname\relax
  \providecommand{\doi}[1]{doi: #1}\else
  \providecommand{\doi}{doi: \begingroup \urlstyle{rm}\Url}\fi

\bibitem[Ahn et~al.(2019)Ahn, Hu, Damianou, Lawrence, and
  Dai]{ahn2019variational}
Sungsoo Ahn, Shell~Xu Hu, Andreas Damianou, Neil~D Lawrence, and Zhenwen Dai.
\newblock Variational information distillation for knowledge transfer.
\newblock In \emph{Proc. IEEE CVPR}, pp.\  9163--9171, 2019.

\bibitem[Ba \& Caruana(2014)Ba and Caruana]{ba2014deep}
Jimmy Ba and Rich Caruana.
\newblock Do deep nets really need to be deep?
\newblock \emph{Proc. NIPS}, 27, 2014.

\bibitem[Bauer \& Kohavi(1999)Bauer and Kohavi]{bauer1999empirical}
Eric Bauer and Ron Kohavi.
\newblock An empirical comparison of voting classification algorithms: Bagging,
  boosting, and variants.
\newblock \emph{Machine learning}, 36\penalty0 (1):\penalty0 105--139, 1999.

\bibitem[Bertinetto et~al.(2016)Bertinetto, Henriques, Valmadre, Torr, and
  Vedaldi]{bertinetto2016learning}
Luca Bertinetto, Jo{\~a}o~F Henriques, Jack Valmadre, Philip Torr, and Andrea
  Vedaldi.
\newblock Learning feed-forward one-shot learners.
\newblock \emph{Proc. NIPS}, 29, 2016.

\bibitem[Bonab \& Can(2019)Bonab and Can]{bonab2019less}
Hamed Bonab and Fazli Can.
\newblock Less is more: A comprehensive framework for the number of components
  of ensemble classifiers.
\newblock \emph{IEEE Transactions on neural networks and learning systems},
  30\penalty0 (9):\penalty0 2735--2745, 2019.

\bibitem[Breiman(1996)]{breiman1996bagging}
Leo Breiman.
\newblock Bagging predictors.
\newblock \emph{Machine learning}, 24\penalty0 (2):\penalty0 123--140, 1996.

\bibitem[Bucilua et~al.(2006)Bucilua, Caruana, and
  Niculescu-Mizil]{bucilua2006model}
Cristian Bucilua, Rich Caruana, and Alexandru Niculescu-Mizil.
\newblock Model compression.
\newblock In \emph{Proc. ACM SIGKDD}, pp.\  535--541, 2006.

\bibitem[Budczies et~al.(2012)Budczies, Klauschen, Sinn, Gy{\H{o}}rffy,
  Schmitt, Darb-Esfahani, and Denkert]{budczies2012cutoff}
Jan Budczies, Frederick Klauschen, Bruno~V Sinn, Bal{\'a}zs Gy{\H{o}}rffy,
  Wolfgang~D Schmitt, Silvia Darb-Esfahani, and Carsten Denkert.
\newblock Cutoff finder: a comprehensive and straightforward web application
  enabling rapid biomarker cutoff optimization.
\newblock \emph{PloS one}, 7\penalty0 (12):\penalty0 e51862, 2012.

\bibitem[Buizza \& Palmer(1998)Buizza and Palmer]{buizza1998impact}
Roberto Buizza and Tim~N Palmer.
\newblock Impact of ensemble size on ensemble prediction.
\newblock \emph{Monthly Weather Review}, 126\penalty0 (9):\penalty0 2503--2518,
  1998.

\bibitem[Caruana et~al.(2004)Caruana, Niculescu-Mizil, Crew, and
  Ksikes]{caruana2004ensemble}
Rich Caruana, Alexandru Niculescu-Mizil, Geoff Crew, and Alex Ksikes.
\newblock Ensemble selection from libraries of models.
\newblock In \emph{Proc. ICML}, pp.\ ~18, 2004.

\bibitem[Caruana et~al.(2006)Caruana, Munson, and
  Niculescu-Mizil]{caruana2006getting}
Rich Caruana, Art Munson, and Alexandru Niculescu-Mizil.
\newblock Getting the most out of ensemble selection.
\newblock In \emph{Proc. ICDM}, pp.\  828--833. IEEE, 2006.

\bibitem[Chen et~al.(2021)Chen, Wang, Gan, Liu, Henao, and
  Carin]{chen2021wasserstein}
Liqun Chen, Dong Wang, Zhe Gan, Jingjing Liu, Ricardo Henao, and Lawrence
  Carin.
\newblock Wasserstein contrastive representation distillation.
\newblock In \emph{Proc. IEEE CVPR}, pp.\  16296--16305, 2021.

\bibitem[Deng et~al.(2009)Deng, Dong, Socher, Li, Li, and
  Fei-Fei]{deng2009imagenet}
Jia Deng, Wei Dong, Richard Socher, Li-Jia Li, Kai Li, and Li~Fei-Fei.
\newblock Imagenet: A large-scale hierarchical image database.
\newblock In \emph{Proc. IEEE CVPR}, pp.\  248--255. Ieee, 2009.

\bibitem[Dietterich(2000)]{dietterich2000ensemble}
Thomas~G Dietterich.
\newblock Ensemble methods in machine learning.
\newblock In \emph{International workshop on multiple classifier systems}, pp.\
   1--15. Springer, 2000.

\bibitem[Dosovitskiy et~al.(2020)Dosovitskiy, Beyer, Kolesnikov, Weissenborn,
  Zhai, Unterthiner, Dehghani, Minderer, Heigold, Gelly,
  et~al.]{dosovitskiy2020image}
Alexey Dosovitskiy, Lucas Beyer, Alexander Kolesnikov, Dirk Weissenborn,
  Xiaohua Zhai, Thomas Unterthiner, Mostafa Dehghani, Matthias Minderer, Georg
  Heigold, Sylvain Gelly, et~al.
\newblock An image is worth 16x16 words: Transformers for image recognition at
  scale.
\newblock In \emph{Proc. ICLR}, 2020.

\bibitem[Drucker et~al.(1994)Drucker, Cortes, Jackel, LeCun, and
  Vapnik]{drucker1994boosting}
Harris Drucker, Corinna Cortes, Lawrence~D Jackel, Yann LeCun, and Vladimir
  Vapnik.
\newblock Boosting and other ensemble methods.
\newblock \emph{Neural Computation}, 6\penalty0 (6):\penalty0 1289--1301, 1994.

\bibitem[Elsken et~al.(2020)Elsken, Staffler, Metzen, and
  Hutter]{elsken2020meta}
Thomas Elsken, Benedikt Staffler, Jan~Hendrik Metzen, and Frank Hutter.
\newblock Meta-learning of neural architectures for few-shot learning.
\newblock In \emph{Proc. IEEE CVPR}, pp.\  12365--12375, 2020.

\bibitem[Freitag et~al.(2017)Freitag, Al-Onaizan, and
  Sankaran]{freitag2017ensemble}
Markus Freitag, Yaser Al-Onaizan, and Baskaran Sankaran.
\newblock Ensemble distillation for neural machine translation.
\newblock \emph{arXiv preprint arXiv:1702.01802}, 2017.

\bibitem[Freund \& Schapire(1997)Freund and Schapire]{freund1997decision}
Yoav Freund and Robert~E Schapire.
\newblock A decision-theoretic generalization of on-line learning and an
  application to boosting.
\newblock \emph{Journal of computer and system sciences}, 55\penalty0
  (1):\penalty0 119--139, 1997.

\bibitem[Gontijo-Lopes et~al.(2021)Gontijo-Lopes, Dauphin, and
  Cubuk]{gontijo2021no}
Raphael Gontijo-Lopes, Yann Dauphin, and Ekin~D Cubuk.
\newblock No one representation to rule them all: Overlapping features of
  training methods.
\newblock \emph{arXiv preprint arXiv:2110.12899}, 2021.

\bibitem[Gou et~al.(2021)Gou, Yu, Maybank, and Tao]{gou2021knowledge}
Jianping Gou, Baosheng Yu, Stephen~J Maybank, and Dacheng Tao.
\newblock Knowledge distillation: A survey.
\newblock \emph{International Journal of Computer Vision}, 129\penalty0
  (6):\penalty0 1789--1819, 2021.

\bibitem[Guo et~al.(2017)Guo, Pleiss, Sun, and Weinberger]{guo2017calibration}
Chuan Guo, Geoff Pleiss, Yu~Sun, and Kilian~Q Weinberger.
\newblock On calibration of modern neural networks.
\newblock In \emph{Proc. ICML}, pp.\  1321--1330. PMLR, 2017.

\bibitem[Ha et~al.(2016)Ha, Dai, and Le]{ha2016hypernetworks}
David Ha, Andrew Dai, and Quoc~V Le.
\newblock Hypernetworks.
\newblock \emph{arXiv preprint arXiv:1609.09106}, 2016.

\bibitem[Hinton et~al.(2015)Hinton, Vinyals, Dean,
  et~al.]{hinton2015distilling}
Geoffrey Hinton, Oriol Vinyals, Jeff Dean, et~al.
\newblock Distilling the knowledge in a neural network.
\newblock \emph{arXiv preprint arXiv:1503.02531}, 2\penalty0 (7), 2015.

\bibitem[Hospedales et~al.(2020)Hospedales, Antoniou, Micaelli, and
  Storkey]{hospedales2020meta}
Timothy Hospedales, Antreas Antoniou, Paul Micaelli, and Amos Storkey.
\newblock Meta-learning in neural networks: A survey.
\newblock \emph{arXiv preprint arXiv:2004.05439}, 2020.

\bibitem[Kim et~al.(2018)Kim, Park, and Kwak]{kim2018paraphrasing}
Jangho Kim, SeongUk Park, and Nojun Kwak.
\newblock Paraphrasing complex network: Network compression via factor
  transfer.
\newblock \emph{Proc. NIPS}, 31, 2018.

\bibitem[Kim \& Rush(2016)Kim and Rush]{kim2016sequence}
Yoon Kim and Alexander~M Rush.
\newblock Sequence-level knowledge distillation.
\newblock In \emph{Proc. EMNLP}, 2016.

\bibitem[Knyazev et~al.(2021)Knyazev, Drozdzal, Taylor, and
  Romero~Soriano]{knyazev2021parameter}
Boris Knyazev, Michal Drozdzal, Graham~W Taylor, and Adriana Romero~Soriano.
\newblock Parameter prediction for unseen deep architectures.
\newblock \emph{Proc. NIPS}, 34, 2021.

\bibitem[Krizhevsky et~al.(2009)Krizhevsky, Hinton,
  et~al.]{krizhevsky2009learning}
Alex Krizhevsky, Geoffrey Hinton, et~al.
\newblock Learning multiple layers of features from tiny images.
\newblock 2009.

\bibitem[Lakshminarayanan et~al.(2017)Lakshminarayanan, Pritzel, and
  Blundell]{lakshminarayanan2017simple}
Balaji Lakshminarayanan, Alexander Pritzel, and Charles Blundell.
\newblock Simple and scalable predictive uncertainty estimation using deep
  ensembles.
\newblock \emph{Proc. NIPS}, 30, 2017.

\bibitem[L{\'e}vesque et~al.(2016)L{\'e}vesque, Gagn{\'e}, and
  Sabourin]{levesque2016bayesian}
Julien-Charles L{\'e}vesque, Christian Gagn{\'e}, and Robert Sabourin.
\newblock Bayesian hyperparameter optimization for ensemble learning.
\newblock \emph{arXiv preprint arXiv:1605.06394}, 2016.

\bibitem[Li et~al.(2019)Li, Dong, Mei, Ma, Huang, and Hu]{li2019lgm}
Huaiyu Li, Weiming Dong, Xing Mei, Chongyang Ma, Feiyue Huang, and Bao-Gang Hu.
\newblock Lgm-net: Learning to generate matching networks for few-shot
  learning.
\newblock In \emph{Proc. ICML}, pp.\  3825--3834. PMLR, 2019.

\bibitem[Lian et~al.(2019)Lian, Zheng, Xu, Lu, Lin, Zhao, Huang, and
  Gao]{lian2019towards}
Dongze Lian, Yin Zheng, Yintao Xu, Yanxiong Lu, Leyu Lin, Peilin Zhao, Junzhou
  Huang, and Shenghua Gao.
\newblock Towards fast adaptation of neural architectures with meta learning.
\newblock In \emph{Proc. ICLR}, 2019.

\bibitem[Lin et~al.(2020)Lin, Kong, Stich, and Jaggi]{lin2020ensemble}
Tao Lin, Lingjing Kong, Sebastian~U Stich, and Martin Jaggi.
\newblock Ensemble distillation for robust model fusion in federated learning.
\newblock \emph{Proc. NIPS}, 33:\penalty0 2351--2363, 2020.

\bibitem[Liu et~al.(2021{\natexlab{a}})Liu, Gao, Zhang, Meng, and
  Lin]{liu2021investigating}
Risheng Liu, Jiaxin Gao, Jin Zhang, Deyu Meng, and Zhouchen Lin.
\newblock Investigating bi-level optimization for learning and vision from a
  unified perspective: A survey and beyond.
\newblock \emph{IEEE Transactions on Pattern Analysis and Machine
  Intelligence}, 2021{\natexlab{a}}.

\bibitem[Liu et~al.(2019)Liu, Cao, Li, Yuan, Hu, Li, and
  Duan]{liu2019knowledge}
Yufan Liu, Jiajiong Cao, Bing Li, Chunfeng Yuan, Weiming Hu, Yangxi Li, and
  Yunqiang Duan.
\newblock Knowledge distillation via instance relationship graph.
\newblock In \emph{Proc. IEEE CVPR}, pp.\  7096--7104, 2019.

\bibitem[Liu et~al.(2021{\natexlab{b}})Liu, Lin, Cao, Hu, Wei, Zhang, Lin, and
  Guo]{liu2021swin}
Ze~Liu, Yutong Lin, Yue Cao, Han Hu, Yixuan Wei, Zheng Zhang, Stephen Lin, and
  Baining Guo.
\newblock Swin transformer: Hierarchical vision transformer using shifted
  windows.
\newblock In \emph{Proc. IEEE CVPR}, pp.\  10012--10022, 2021{\natexlab{b}}.

\bibitem[Mahabadi et~al.(2021)Mahabadi, Ruder, Dehghani, and
  Henderson]{mahabadi2021parameter}
Rabeeh~Karimi Mahabadi, Sebastian Ruder, Mostafa Dehghani, and James Henderson.
\newblock Parameter-efficient multi-task fine-tuning for transformers via
  shared hypernetworks.
\newblock In \emph{Proc. ACL}, pp.\  565--576, 2021.

\bibitem[Malinin et~al.(2019)Malinin, Mlodozeniec, and
  Gales]{malinin2019ensemble}
Andrey Malinin, Bruno Mlodozeniec, and Mark Gales.
\newblock Ensemble distribution distillation.
\newblock In \emph{Proc. ICLR}, 2019.

\bibitem[Mendoza et~al.(2016)Mendoza, Klein, Feurer, Springenberg, and
  Hutter]{mendoza2016towards}
Hector Mendoza, Aaron Klein, Matthias Feurer, Jost~Tobias Springenberg, and
  Frank Hutter.
\newblock Towards automatically-tuned neural networks.
\newblock In \emph{Workshop on Automatic Machine Learning}, pp.\  58--65. PMLR,
  2016.

\bibitem[Opitz \& Maclin(1999)Opitz and Maclin]{opitz1999popular}
David Opitz and Richard Maclin.
\newblock Popular ensemble methods: An empirical study.
\newblock \emph{Journal of artificial intelligence research}, 11:\penalty0
  169--198, 1999.

\bibitem[Ovadia et~al.(2019)Ovadia, Fertig, Ren, Nado, Sculley, Nowozin,
  Dillon, Lakshminarayanan, and Snoek]{ovadia2019can}
Yaniv Ovadia, Emily Fertig, Jie Ren, Zachary Nado, David Sculley, Sebastian
  Nowozin, Joshua Dillon, Balaji Lakshminarayanan, and Jasper Snoek.
\newblock Can you trust your model's uncertainty? evaluating predictive
  uncertainty under dataset shift.
\newblock \emph{Proc. NIPS}, 32, 2019.

\bibitem[Park et~al.(2021)Park, Cha, Kim, Han, et~al.]{park2021learning}
Dae~Young Park, Moon-Hyun Cha, Daesin Kim, Bohyung Han, et~al.
\newblock Learning student-friendly teacher networks for knowledge
  distillation.
\newblock \emph{Proc. NIPS}, 34:\penalty0 13292--13303, 2021.

\bibitem[Park et~al.(2019)Park, Kim, Lu, and Cho]{park2019relational}
Wonpyo Park, Dongju Kim, Yan Lu, and Minsu Cho.
\newblock Relational knowledge distillation.
\newblock In \emph{Proc. IEEE CVPR}, pp.\  3967--3976, 2019.

\bibitem[Peng et~al.(2019)Peng, Jin, Liu, Li, Wu, Liu, Zhou, and
  Zhang]{peng2019correlation}
Baoyun Peng, Xiao Jin, Jiaheng Liu, Dongsheng Li, Yichao Wu, Yu~Liu, Shunfeng
  Zhou, and Zhaoning Zhang.
\newblock Correlation congruence for knowledge distillation.
\newblock In \emph{Proc. IEEE ICCV}, pp.\  5007--5016, 2019.

\bibitem[Perrone \& Cooper(1992)Perrone and Cooper]{perrone1992networks}
Michael~P Perrone and Leon~N Cooper.
\newblock When networks disagree: Ensemble methods for hybrid neural networks.
\newblock Technical report, Brown Univ Providence Ri Inst for Brain and Neural
  Systems, 1992.

\bibitem[Polino et~al.(2018)Polino, Pascanu, and Alistarh]{polino2018model}
Antonio Polino, Razvan Pascanu, and Dan Alistarh.
\newblock Model compression via distillation and quantization.
\newblock In \emph{Proc. ICLR}, 2018.

\bibitem[Ratzlaff \& Fuxin(2019)Ratzlaff and Fuxin]{ratzlaff2019hypergan}
Neale Ratzlaff and Li~Fuxin.
\newblock Hypergan: A generative model for diverse, performant neural networks.
\newblock In \emph{Proc. ICML}, pp.\  5361--5369. PMLR, 2019.

\bibitem[Rokach(2010)]{rokach2010ensemble}
Lior Rokach.
\newblock Ensemble-based classifiers.
\newblock \emph{Artificial intelligence review}, 33\penalty0 (1):\penalty0
  1--39, 2010.

\bibitem[Rusu et~al.(2018)Rusu, Rao, Sygnowski, Vinyals, Pascanu, Osindero, and
  Hadsell]{rusu2018meta}
Andrei~A Rusu, Dushyant Rao, Jakub Sygnowski, Oriol Vinyals, Razvan Pascanu,
  Simon Osindero, and Raia Hadsell.
\newblock Meta-learning with latent embedding optimization.
\newblock In \emph{Proc. ICLR}, 2018.

\bibitem[Sagi \& Rokach(2018)Sagi and Rokach]{sagi2018ensemble}
Omer Sagi and Lior Rokach.
\newblock Ensemble learning: A survey.
\newblock \emph{Wiley Interdisciplinary Reviews: Data Mining and Knowledge
  Discovery}, 8\penalty0 (4):\penalty0 e1249, 2018.

\bibitem[Saikia et~al.(2020)Saikia, Brox, and Schmid]{saikia2020optimized}
Tonmoy Saikia, Thomas Brox, and Cordelia Schmid.
\newblock Optimized generic feature learning for few-shot classification across
  domains.
\newblock \emph{arXiv preprint arXiv:2001.07926}, 2020.

\bibitem[Snell et~al.(2017)Snell, Swersky, and Zemel]{snell2017prototypical}
Jake Snell, Kevin Swersky, and Richard Zemel.
\newblock Prototypical networks for few-shot learning.
\newblock \emph{Proc. NIPS}, 30, 2017.

\bibitem[Tay et~al.(2020)Tay, Zhao, Bahri, Metzler, and Juan]{tay2020hypergrid}
Yi~Tay, Zhe Zhao, Dara Bahri, Donald Metzler, and Da-Cheng Juan.
\newblock Hypergrid transformers: Towards a single model for multiple tasks.
\newblock In \emph{Proc. ICLR}, 2020.

\bibitem[Tian et~al.(2019)Tian, Krishnan, and Isola]{tian2019contrastive}
Yonglong Tian, Dilip Krishnan, and Phillip Isola.
\newblock Contrastive representation distillation.
\newblock In \emph{Proc. ICLR}, 2019.

\bibitem[Tung \& Mori(2019)Tung and Mori]{tung2019similarity}
Frederick Tung and Greg Mori.
\newblock Similarity-preserving knowledge distillation.
\newblock In \emph{Proc. IEEE CVPR}, pp.\  1365--1374, 2019.

\bibitem[Vaswani et~al.(2017)Vaswani, Shazeer, Parmar, Uszkoreit, Jones, Gomez,
  Kaiser, and Polosukhin]{attention}
Ashish Vaswani, Noam Shazeer, Niki Parmar, Jakob Uszkoreit, Llion Jones,
  Aidan~N. Gomez, Lukasz Kaiser, and Illia Polosukhin.
\newblock Attention is all you need.
\newblock In \emph{Proc. NIPS}, pp.\  5998--6008, 2017.

\bibitem[Wang \& Yoon(2021)Wang and Yoon]{wang2021knowledge}
Lin Wang and Kuk-Jin Yoon.
\newblock Knowledge distillation and student-teacher learning for visual
  intelligence: A review and new outlooks.
\newblock \emph{IEEE Transactions on Pattern Analysis and Machine
  Intelligence}, 2021.

\bibitem[Wenzel et~al.(2020)Wenzel, Snoek, Tran, and
  Jenatton]{wenzel2020hyperparameter}
Florian Wenzel, Jasper Snoek, Dustin Tran, and Rodolphe Jenatton.
\newblock Hyperparameter ensembles for robustness and uncertainty
  quantification.
\newblock \emph{Proc. NIPS}, pp.\  6514--6527, 2020.

\bibitem[Wortsman et~al.(2022)Wortsman, Ilharco, Gadre, Roelofs, Gontijo-Lopes,
  Morcos, Namkoong, Farhadi, Carmon, Kornblith, et~al.]{wortsman2022model}
Mitchell Wortsman, Gabriel Ilharco, Samir~Ya Gadre, Rebecca Roelofs, Raphael
  Gontijo-Lopes, Ari~S Morcos, Hongseok Namkoong, Ali Farhadi, Yair Carmon,
  Simon Kornblith, et~al.
\newblock Model soups: averaging weights of multiple fine-tuned models improves
  accuracy without increasing inference time.
\newblock In \emph{Proc. ICML}, pp.\  23965--23998. PMLR, 2022.

\bibitem[Ye \& Ren(2021)Ye and Ren]{ye2021learning}
Qinyuan Ye and Xiang Ren.
\newblock Learning to generate task-specific adapters from task description.
\newblock In \emph{Proc. ACL}, pp.\  646--653, 2021.

\bibitem[Zhao et~al.(2022)Zhao, Cui, Song, Qiu, and Liang]{zhao2022decoupled}
Borui Zhao, Quan Cui, Renjie Song, Yiyu Qiu, and Jiajun Liang.
\newblock Decoupled knowledge distillation.
\newblock In \emph{Proc. IEEE CVPR}, pp.\  11953--11962, 2022.

\bibitem[Zhmoginov et~al.(2022)Zhmoginov, Sandler, and
  Vladymyrov]{zhmoginov2022}
Andrey Zhmoginov, Mark Sandler, and Maksym Vladymyrov.
\newblock Hypertransformer: Model generation for supervised and semi-supervised
  few-shot learning.
\newblock In \emph{Proc. ICML}, pp.\  27075--27098. PMLR, 2022.

\end{thebibliography}
\bibliographystyle{iclr2023_conference}



\end{document}